\documentclass[sigconf]{aamas} 
\usepackage{balance} % for balancing columns on the final page

\usepackage[utf8]{inputenc} % allow utf-8 input
\usepackage[T1]{fontenc}    % use 8-bit T1 fonts
\usepackage{hyperref}       % hyperlinks
\usepackage{url}            % simple URL typesetting
\usepackage{booktabs}       % professional-quality tables
\usepackage{amsfonts}       % blackboard math symbols
\usepackage{nicefrac}       % compact symbols for 1/2, etc.
\usepackage{microtype}      % microtypography
\usepackage{lipsum}		% Can be removed after putting your text content
\usepackage{graphicx}
\usepackage{natbib}
\usepackage{doi}
\usepackage{soul, xcolor}
\usepackage{amsmath}%, amssymb}
\usepackage{MnSymbol}
\usepackage{relsize}
\usepackage{caption}
\usepackage{subcaption}

\setcopyright{ifaamas}
\acmConference[AAMAS '24]{Proc.\@ of the 23rd International Conference
on Autonomous Agents and Multiagent Systems (AAMAS 2024)}{May 6 -- 10, 2024}
{Auckland, New Zealand}{N.~Alechina, V.~Dignum, M.~Dastani, J.S.~Sichman (eds.)}
\copyrightyear{2024}
\acmYear{2024}
\acmDOI{}
\acmPrice{}
\acmISBN{}

\acmSubmissionID{588}

\title[Towards Scalable and Efficient Causal Discovery with Reinforcement Learning]{CORE: Towards Scalable and Efficient Causal Discovery with Reinforcement Learning}

\author{Andreas Sauter}
\affiliation{
  \institution{Vrije Universiteit Amsterdam}
  \city{Amsterdam}
  \country{The Netherlands}}
\email{a.sauter@vu.nl}

\author{Nicolò Botteghi}
\affiliation{
  \institution{University of Twente}
  \city{Enschede}
  \country{The Netherlands}}
\email{n.botteghi@utwente.nl}

\author{Erman Acar}
\affiliation{
  \institution{IvI and ILLC, University of Amsterdam }
  \city{Amsterdam}
  \country{The Netherlands}}
\email{e.acar@uva.nl}

\author{Aske Plaat}
\affiliation{
  \institution{LIACS, Leiden University}
  \city{Leiden}
  \country{The Netherlands}}
\email{
a.plaat@liacs.leidenuniv.nl}

\begin{abstract}
Causal discovery is the challenging task of inferring causal structure from data. Motivated by Pearl’s Causal Hierarchy (PCH), which tells us that passive observations alone are not enough to distinguish correlation from causation, there has been a recent push to incorporate interventions into machine learning research. Reinforcement learning provides a convenient framework for such an active approach to learning. This paper presents CORE, a deep reinforcement learning-based approach for causal discovery and intervention planning. CORE learns to sequentially reconstruct causal graphs from data while learning to perform informative interventions. Our results demonstrate that CORE generalizes to unseen graphs and efficiently uncovers causal structures. Furthermore, CORE scales to larger graphs with up to 10 variables and outperforms existing approaches in structure estimation accuracy and sample efficiency. All relevant code and supplementary material can be found at \url{https://github.com/sa-and/CORE}. %CORE: Causal discOvery Reinforcement lEarning
\end{abstract} 


\newcommand{\BibTeX}{\rm B\kern-.05em{\sc i\kern-.025em b}\kern-.08em\TeX}

\begin{document}

%%% The following commands remove the headers in your paper. For final 
%%% papers, these will be inserted during the pagination process.

\pagestyle{fancy}
\fancyhead{}

%%% The next command prints the information defined in the preamble.

\maketitle 

%\def\thefootnote{*}\footnotetext{These authors contributed equally to this work.}

%%%%%%%%%%%%%%%%%%%%%%%%%%%%%%%%%%%%%%%%%%%%%%%%%%%%%%%%%%%%%%%%%%%%%%%%


\section{Introduction and Related Work}
 %is 
% a challenging endeavor dedicated to 
Causal discovery (CD) is the challenging task of inferring causal structure from data~\cite{Glymour19Review, Vowels23yall}.  Traditional approaches to causal discovery consider data from purely observational distributions. These are approaches such as constraint-based ones \cite{Spirtes2000CausationSearch,Hyttinen2014Constraint-basedProgramming}, score-based ones \cite{Chickering2003OptimalSearch}, and more recently continuous optimization-based ones \cite{Zheng2018DAGsLearning,Yu2019DAG-GNN:Networks}. 

Pearl's Causal Hierarchy (PCH) asserts that distinguishing between mere correlations and genuine causal relationships requires the integration of interventions in general \cite{Bareinboim2022OnInference}.  As a response to this requirement, there has been a recent push to incorporate interventions into causal discovery research \cite{Mooij2020JointContexts, Hauser2012CharacterizationUhlmann} including machine learning \cite{Brouillard2020DifferentiableData,Lippe2021EfficientConstraints,Sauter2023ADiscovery}, among others. 

Reinforcement learning (RL) learns an optimal policy for sequential decision problems through interactions \cite{Sutton2018ReinforcementIntroduction}. Therefore, RL is a promising framework for using interventions to investigate causal relationships. In particular, RL plays a dual role in the realm of causal discovery - it can be used not only to recover the causal structure of an environment \cite{Zhu2019CausalLearning}, but also to learn causal discovery algorithms \cite{Sauter2023ADiscovery}, thus representing a versatile tool for CD. 

In particular, RL has also been used to search the space of causal structures more efficiently based on a fixed dataset \cite{Zhu2019CausalLearning, Wang21CORL} with the possibility of incorporating prior knowledge \citep{Hasan2022Kcrl:Learning}. Similarly, work on RL-related GFlow Nets \citep{Bengio2023GFlowNetFoundations} has been deployed to generate good estimates of the true causal structure \citep{Deleu2022BayesianNetworks, Li2022GFlowCausal:Discovery}. Furthermore, many integrations of RL with causal concepts have been investigated that restrict their CD process to supervised learning \citep{Nair2019CausalTasks, Ton2021MetaDirection, Mendez-Molina2022CausalIntegration, Lei2022CausalModels, Feng2022FactoredLearning}.  In addition to that, RL has also been used to learn policies that choose the best interventions to do for CD \citep{Amirinezhad2022ActiveLearning, Scherrer2022LearningInterventions, Tigas2022InterventionsScale}. 

Although causal discovery has seen substantial progress with these works over the years, leading to a multitude of methodologies, challenges persist in areas such as scalability, generalization, and planning of interventions. In this context, this paper introduces CORE (\textbf{C}ausal Disc\textbf{O}very with \textbf{RE}inforcement Learning), a deep-RL-based algorithm designed for the task of learning a CD policy. CORE can learn a policy that sequentially reconstructs causal graphs from both observational and interventional data, while simultaneously performing informative interventions. This dual learning paradigm allows CORE not only to uncover causal structures efficiently, but also to identify interventions that enhance its causal models. The following lists our main contributions:
\begin{itemize}
    \item We formalize the task of learning a CD algorithm as a partially observable Markov decision process (POMDP).
    \item We propose a dual Q-learning setup to learn intervention design and structure estimation simultaneously.
    \item We demonstrate that CORE can be successfully applied for causal discovery to previously unseen graphs of sizes of up to 10 variables. 
\end{itemize}
In addition to those, we show the importance of jointly learning which interventions to perform and graph generation, and investigate the limitations of our approach regarding the applicability to the real world.

The most distinctive feature of CORE is that it does not impose a specific algorithm for identifying causal models, but rather attempts to learn it. Among others, this can have positive effects on efficiency and transferability to new problem instances. While MCD \cite{Sauter2023ADiscovery} and AVICI \cite{Lorch2022AmortizedLearning} solve the same task, they run into pitfalls that hinder their application to realistic graph sizes or rely on offline data, respectively. We set steps to overcome these pitfalls by imposing additional structure on our policy, more efficient rewards, and learning to actively perform relevant interventions.

Our results show robust generalization to unseen graphs and the capability to scale to scenarios with up to ten variables, a step forward over the state of the art, and a crucial advancement towards addressing real-world complexities. %In terms of structure estimation accuracy, CORE also surpasses existing methodologies. 
The subsequent sections delve into the intricacies of CORE's architecture, its training methodologies, and empirical validations.
%, which collectively contribute to the field of causal discovery.

%\textcolor{red}{Reinforcement Learning for Causal Discovery (sequential construction of the graph vs in one go)}
%An appealing alternative to directly inferring the whole causal graph in one forward step is the use of deep reinforcement learning \hl{add ref} (DRL). DRL studies the problem of optimal sequential decision-making of an agent that interacts with an unknown environment through its action. The agent's goal is to find the sequence of actions maximizing the reward function, i.e. the indicator of the task's optimality. In the context of causal discovery, DRL can be used to sequentially learn to construct the causal graph by adding or removing 1 edge at each timestep.

%\textcolor{red}{Intervention Design}
%Often approaches come with predefined interventions, but there is also the question of which intervention to perform.

%\textcolor{red}{Contributions of our paper}
%With reference to Figure \hl{1}, we introduce a DRL-based causal discovery algorithm composed of two agents. The first agent, i.e. the interventional agent, learns to perform optimal intervention by interacting with the environment, i.e. the scm \hl{add more details here}. The second agent instead aims to learn the causal graph using interventional and observational data. We frame this problem in the context of meta learning \hl{add ref}, where our goal is to train our agents on a set of different causal graph and eventually develop a causal discovery algorithm that can generalize to unknown ones.

\section{Preliminaries and Notation}
In this section, we establish the necessary notation and provide an overview of key concepts and techniques used in the field of causal discovery with interventions and reinforcement learning.

\subsection{Causal Models}\label{subsec:CausalModels}
%Causal discovery relies on the fundamental notion of \emph{causal graphs}. As a common assumption in the field, a causal graph does not contain feedback loops, hence it does not contain cycles. The nodes of a causal graph represent the random variables that describe the system at hand.  Formally, a causal graph is defined as a directed acyclic graph (DAG) $G = (\mathcal{V}, \mathcal{E})$, where $\mathcal{V}$ are random variables and $\mathcal{E}$ are the edges of the graph.  A directed edge between two nodes in a causal graph describes a direct causal relationship between the two. 

Causal relations are often formalized through a \emph{structural causal model} (SCM) which is a tuple $M = (\mathcal{X}, \mathcal{F}, \mathcal{U}, \mathcal{P})$ with a set of endogenous (random) variables (i.e., relevant variables for the problem) $\mathcal{X} = \{X_1, \ldots, X_n\}$,  $\mathcal{U} = \{U_1, \ldots, U_n\}$ a set of exogenous (random) variables (often also called unobservable or noise variables),  $\mathcal{F} =\{f_1, \ldots, f_n\}$ the set of functions (also called \emph{structural equations}) whose elements are in the form of $X_i \leftarrow f_i(Pa(X_i), U_i)$ where $Pa(X_i) \subseteq \mathcal{X} \backslash \{X_i\}$ stands for endogenous parent variables of $X_i$, and $\mathcal{P} = \{P_1, ..., P_n\}$ the set of pairwise independent probability distributions defined over $\mathcal{U}$ with $U_i \sim P_i$. 

Interpreting variables as nodes and the functional dependency between variables as directed edges, every SCM $M$ induces a directed graph structure $G$, which we will call the corresponding causal graph.  Directed edges represent direct causation from parent nodes to child nodes, hence absence of edges is as important as present edges. For the sake of simplicity, we shall follow the common assumption that no variable is its own cause i.e., there is no circular functional dependency, hence the induced causal graph is always a directed acyclic graph (DAG).  Furthermore, each SCM $M$ induces a joint distribution $P_M(\mathcal{X})$ over its endogenous variables, whose structural properties inherited from the corresponding induced graph $G$ satisfy the Markov condition. That is, each $X_i$ is independent of its non-descendants, given its parents $Pa(X_i$). Along with the independence of the noise variables, this condition implies the following  factorization \cite{Pearl2009IntroductionModels}:

%Whenever it is clear from the context, we will ignore the subscript $G$. As per common assumption in the field, our SCMs do not contain feedback loops, hence their induced graphs do not contain cycles and are described as \textit{directed acyclic graphs} (DAGs). 

\begin{equation}
    P_M(\mathcal{X}) =  \prod_{X_i \in \mathcal{X}}P(X_i | Pa(X_i))
\end{equation}
We shall refer to this distribution as \emph{observational distribution}.

Note that SCMs are generative models, i.e., we can sample values for $\mathcal{X}$ from them. We can sample the exogenous variables from $\mathcal{P}$ and determine the values of endogenous variables according to their functions in $\mathcal{F}$. This procedure effectively corresponds to sampling from the joint distribution over endogenous variables \cite{Pearl2009IntroductionModels}.

%A slightly more coarse-grained description of causal relations is by means of \emph{causal Bayesian networks} (BN) that describe the distributions of the variables rather than their functional relation. A BN $B = (P, G)$ is a joint distribution $P(\mathcal{V})$  over variables $V$ that are factorized according to $G$ as $P(\mathcal{V}) = \prod_{i=0}^{\mid \mathcal{V} \mid}P(V_i | Pa_G(V_i))$. Such a network $B$ is causal if $G$ is the causal graph of the variables. 

\subsection{Interventions}\label{subsec:interventions}
Interventions play a crucial role in causal discovery, allowing us to investigate causal relationships by actively manipulating variables in a system. In general, imposed by Pearl's causal hierarchy \cite{Bareinboim2022OnInference}, interventions are necessary to distinguish causation from correlation, and eventually to reason about causal effects. 

%Formally, an \emph{intervention} on a variable $X$ changes the variable's value to $x$, independently of $X$'s actual causes. This operation is the so-called \emph{do-operation} (denoted as $do(X = x)$), and it allows us to distinguish the causal effect of variable $X$ on a set of variables $\mathcal{Y}$ from the influence of $Pa(X)$ on $\mathcal{Y}$ caused by pathways in the causal graph involving $X$ (see Figure \ref{fig:intervention}).

\begin{figure}
    \centering
    \includegraphics[width=0.49\textwidth]{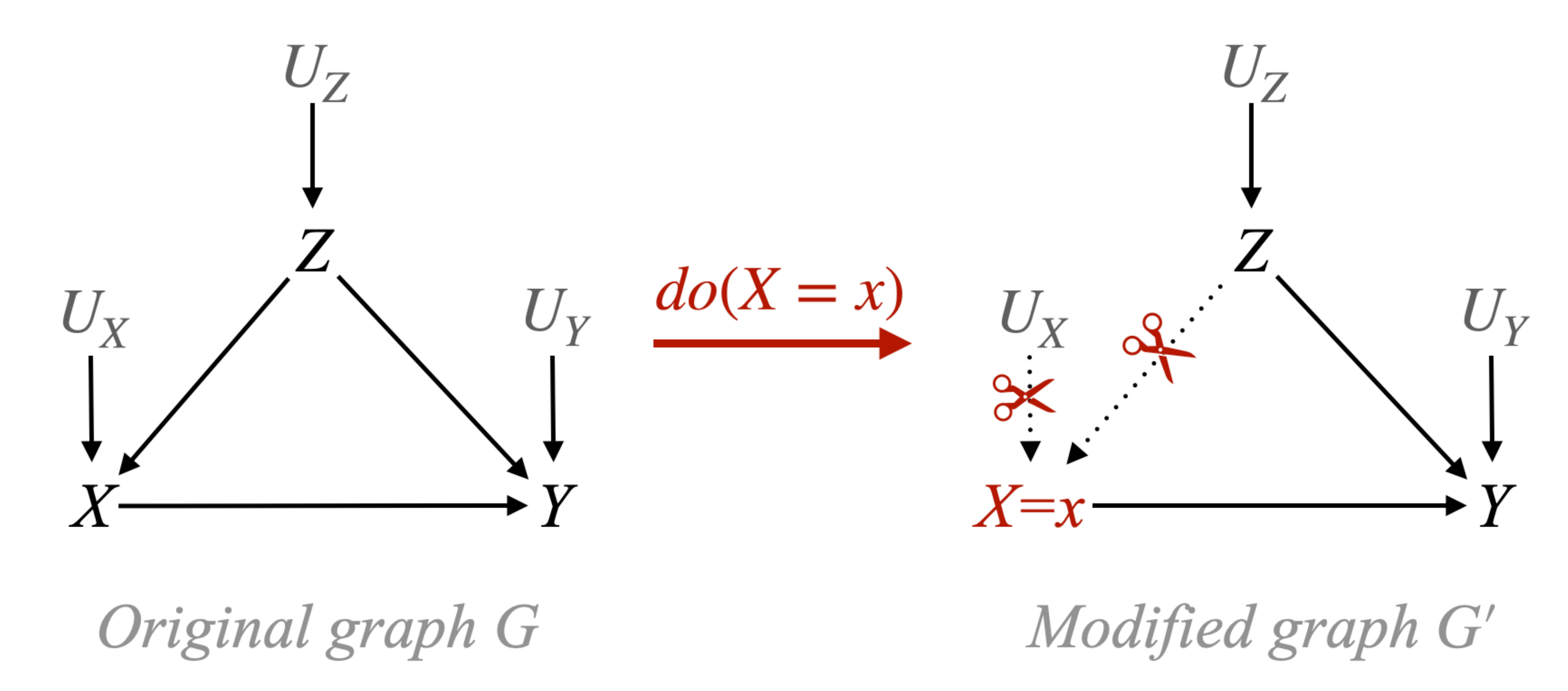}
    \caption{A simple graphical illustration of a (hard) intervention. Given the causal graph $G$ with endogenous variables $\mathcal{X}=\{X, Y, Z\}$ and the corresponding noise variables $\mathcal{U}=\{U_X, U_Y, U_Z\}$, intervening on variable $X$ (i.e., $do(X=x)$) results in modifying $G$ into $G'$ by pruning the incoming edges to node $X$ and assigning the value $x$.
    %This figure demonstrates the effects of an intervention. Red arrows represent causal paths from $\mathcal{Z}$ to $\mathcal{Y}$, and the green arrow signifies the direct effect of $X$ on $\mathcal{Y}$. On the left is the original graph, including a causal path from $\mathcal{Z}$ to $\mathcal{Y}$ through $X$. On the right is the post-intervention graph, showing only the direct causal connection between $X$ and $\mathcal{Y}$."
    }
    \label{fig:intervention}
\end{figure}

Formally, an \emph{intervention} on a variable $X$ changes the variable's value to $x$ (an arbitrary but fixed value), independently of $X$'s actual causes. Then $X$ is called the \emph{intervention target}. Effectively, at the graph level, intervening on a variable $X$, removes all the edges incoming to $X$, resulting in $Pa(X) = \emptyset$. This operation is the so-called \emph{do-operation} (denoted as $do(X = x)$), and allows us to distinguish the causal effect of variable $X$ on variable(s) $Y$ from the confounding influence of common parents of $X$ and $Y$ (Figure \ref{fig:intervention}). 

 In an SCM, intervening on a variable $X$ implies that the corresponding structural equation $f_X \in \mathcal{F}$ is replaced by $X \leftarrow x$, resulting in a modified SCM $M'$. Therefore, an intervention affects the distribution of the intervention target, since: 
\begin{equation}
    P_{M'}(X|Pa(X))=P_{M'}(X|\emptyset)=P_{M'}(X) = \delta_{x}
\end{equation}
where $\delta_{x}$ is the probability density function that has all mass on $x$. Put differently, an intervention replaces the factor associated with the intervened variable. We refer to the resulting joint distribution
%\begin{equation}
%    P'(\mathcal{X}|do(X=x)) = \prod_{X_i \in \mathcal{X} \setminus \{X\}}P(X_i | Pa(X_i))\times P'(X)
%\end{equation} 
\begin{equation}
    P_M(\mathcal{X}|do(X=x)) = \prod_{X_i \in \mathcal{X} \setminus \{X\}}P_{M}(X_i | Pa(X_i))\cdot P_{M'}(X=x)
\end{equation} 
as \emph{post-interventional} distribution. To simplify the notation, we will sometimes use $P_{M_{do(X=x)}}(\mathcal{X})$ or $P_{M_{do(X)}}(\mathcal{X})$ to refer to the expression $P_M(\mathcal{X}|do(X=x))$ when the target variable or $x$ is clear from the context.

%Both SCMs and BNs are generative model i.e. we can sample values for $\mathcal{V}$ from them. For SCMs this can be done by sampling the exogenous variables from $\mathcal{P}$ and determining the values of the endogenous variables $\mathcal{X}$ according to their functions in $\mathcal{F}$. For BNs we can simply sample from the joint distribution $P(\mathcal{V})$. If the models are not intervened on we call the distribution of these samples the \emph{observational distribution} which we denote as $P(\mathcal{V})$. If they are intervened on we call the distribution of the samples \emph{interventional distribution}. For a set of intervened variables $\mathcal{I}$ and corresponding values $\mathbf{i}$, we denote the interventional distribution distribution as $P(\mathcal{V}|do(\mathcal{I}=\mathbf{i}))$.

\subsection{Reinforcement Learning}\label{subsec:RL}
Reinforcement learning (RL) is a general approach to learning through interaction with the world \cite{Sutton2018ReinforcementIntroduction}, especially in sequential decision problems. An RL agent aims to find the sequence of actions that maximize the expected \textit{return}, i.e., the cumulative (discounted) \emph{reward}. How the RL agent selects its actions only relies on (possibly indirect) measures of how the world changes after each action is taken.

%\paragraph{MDPs:} A reinforcement learning problem can be specified formally as a \emph{Markov Decision Processes (MDP)}. An MDP is a tuple $(\mathcal{S},\mathcal{A}, T, R)$ where $\mathcal{S}$ is the set of states, $\mathcal{A}$ is the set of actions, $T: \mathcal{S}\times\mathcal{A}\longrightarrow\mathcal{S}$ \st{$T: \mathcal{S}\times\mathcal{A}\times\mathcal{S} \longrightarrow [0, 1]$} is the \hl{deterministic} transition function characterizing the transition dynamics of the environment, i.e. \hl{the next state}\st{the probability of visiting state} $s'$ when being in state $s$ and taking action $a$, and $R:\mathcal{S}\times\mathcal{A} \longrightarrow \mathbb{R}$ is the reward function that the agent seeks to maximize \cite{Sutton2018ReinforcementIntroduction}. 

\paragraph{POMDP} In reinforcement learning problems, the relationship between an agent and the environment in which the states are not fully observable is often modeled as a Partially Observable Markov Decision Process (POMDP).  Formally, a POMDP is a tuple $(\mathcal{S, A, T, R}, \Omega, \mathcal{O}, \gamma)$ where $\mathcal{S}$ is a set of states, $\mathcal{A}$ is a set of actions, $\mathcal{T}: \mathcal{S}\times\mathcal{A} \times \mathcal{S} \longrightarrow [0, 1]$ is a set of transition probabilities between states, $\mathcal{R}: \mathcal{S} \times \mathcal{A} \rightarrow \mathbb{R}$ is the reward function, $\Omega$ is a set of observations,  $\mathcal{O}: \mathcal{S} \times \mathcal{A} \times \Omega \rightarrow [0, 1]$ is the set of conditional observation probabilities, and $\gamma \in [0, 1)$ is the discount factor.

\paragraph{RL:} The  strategy of actions of an RL agent is called \emph{policy}. A policy can be either deterministic or stochastic. A deterministic policy $\pi:\mathcal{S}\longrightarrow\mathcal{A}$ maps states to an action, whereas a stochastic policy $\pi:\mathcal{S} \times\mathcal{A} \longrightarrow [0, 1]$ is characterized by a conditional distribution of actions given states. To maximize the return in the long run, RL agents often estimate the so-called \emph{value} function $V:\mathcal{S}\longrightarrow\mathbb{R}$ or \emph{ action value} function $Q:\mathcal{S}\times\mathcal{A}\longrightarrow\mathbb{R}$. These functions determine the desirability of a state or a state-action pair, respectively. The optimal Q-function $Q^*$ allows us to derive an optimal policy $\pi^*$ that maximizes the return by greedily choosing the action that maximizes the value of each state, that is, $\pi^*(s) = \mathrm{argmax}_a Q^*(s, a)$ \cite{Sutton2018ReinforcementIntroduction}.

\paragraph{Deep Q-Learning:} The Q-learning algorithm \cite{Watkins1992Q-Learning} estimates the state-action value function $Q$ using the \textit{ temporal difference} (TD). In particular, TD learning decomposes the problem of estimating the expected return of a given policy as the sum of the instantaneous reward and the value accumulated by following the optimal policy in the next step:
\begin{equation}
    Q(s, a) = r(s, a) + \gamma \mathrm{max}_{a'} Q(s', a')
\end{equation}
where $r(s,a)$ is the instantaneous reward of the state-action pair. To estimate the future reward, we assume that the agent follows the optimal policy $\pi^*(s)$, i.e., $\mathrm{argmax}_a Q^*(s, a)$. Especially in the first iterations of the algorithm, the estimate of $Q$ does not correspond to the optimal value function $Q^*$. However, tabular Q-learning can still converge to the optimal solution \cite{Watkins1992Q-Learning}. 
The \emph{deep Q-network (DQN)} algorithm \cite{Mnih2015Human-levelLearning} adapts the Q-learning algorithm to non-tabular settings, e.g., continuous state spaces, where the Q-function needs to be approximated via a neural network. DQN utilizes the TD-learning rule to generate a target for the training of the neural network that approximates the Q-value by means of the loss function:
\begin{equation}
    \mathcal{L}(\theta) = \mathbb{E}_{s,a,r,s'}[(r(s, a) + \gamma \mathrm{max}_{a'} Q(s', a'; \theta^-) - Q(s, a; \theta) )^2 ]
\end{equation}
where $Q(s, a; \theta)$ is the Q-function approximated by the neural network of parameters $\theta$, while $Q(s, a; \theta^-)$ is the so-called target network, used to generate a fixed target and stabilize the training dynamics, and $\gamma$ is the discount factor. 

% \st{The central idea to \emph{Q-learning} is the observation that, if the Q function is optimal, then $Q^*(s, a) = r(s', a) + \mathrm{argmax}_a Q^*(s', a) $. This states that the value of the current state-action pair $(s, a)$ equals the maximum value of the Q function in the next state $s'$ added to the immediate reward $r(s',a)$ \cite{Sutton2018ReinforcementIntroduction}. The \emph{deep Q-network (DQN)} algorithm \cite{Mnih2015Human-levelLearning} adapts this algorithm to make it feasible in a non-tabular setting where the Q function needs to be approximated via a neural network.  }

%\paragraph{The MDP Formulation is not Enough for Causal Discovery:} The problem of causal discovery and optimal interventional design cannot be modeled as an MDP as the state of the environment is not directly observable by the agent. In Section \ref{sec:formulation}, we will show how to cast this problem into a \textit{partially observable} MDP (POMDP).

\section{Learning a Causal Discovery Policy with Informative Interventions}%Reinforcement Learning of Causal Structures through Observational and Interventional Data}
\label{sec:setup} %methodology 
\begin{figure*}[t]
    \centering
    \includegraphics[width=\textwidth]{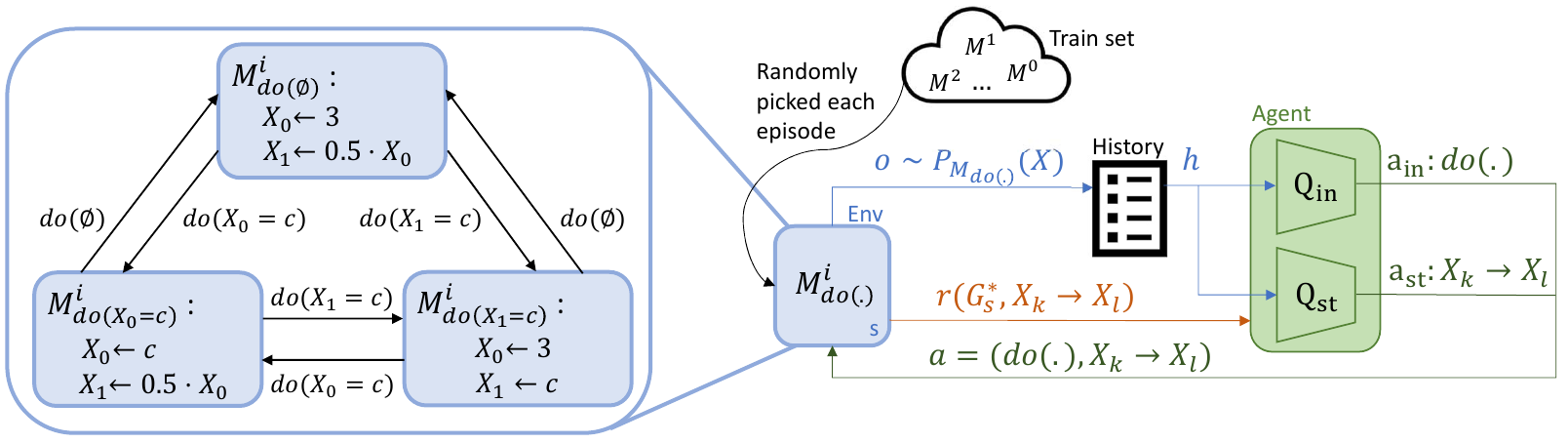}
    \caption{Overview of COREs training setup (right) and a minimal example of the transition dynamics for an SCM with two endogenous variables (left). At each step, the agent picks the intervention/structural actions according to an $\epsilon$-greedy policy on $Q_{in}$ and $Q_{st}$ respectively. The intervention is applied to the SCM $M^i$ leading to a post-interventional distribution $P_{M_{do(.)}}$ from which an observation is sampled. The agent receives a reward based on the structure action and the induced graph of $M_{do(\emptyset)}^i$. The observation is added to the history of observations and serves as input to the agent. At the beginning of each episode a new $M^j$ is drawn from the training set and the observation history is cleared.}
    \label{fig:overview}
\end{figure*}
In this section, we present our algorithmic setup for learning causal discovery policies. We consider the classic agent-environment interaction scheme commonly used in RL. The goal is to learn a policy that represents a CD algorithm that uses observational and interventional data to sequentially estimate the true causal structures by performing informative interventions. Such modules can be applied to previously unseen causal structures through a few forward passes of a neural network without retraining, making them a highly efficient tool for causality \cite{Lorch2022AmortizedLearning, Sauter2023ADiscovery}.

Following this line of research, we learn a policy that collects a stream of data by intervening on the environment to infer a causal structure estimate. This setup acknowledges the strong influence that informative interventions have, especially when there is a limited budget for interventions. We learn to perform these interventions by rewarding interventions that lead to the generation of a better structural update and limiting the budget for interventions by means of possible steps in an episode. 

One key aspect of learned CD modules is that they need information about the ground truth causal structure only during training. This promises the possibility of (i) training the CD policy on synthetic data where the ground truth can easily be generated, and then (ii) applying it to estimate the causal structure of environments where the ground truth structure is potentially unknown, such as in the real world. %As we shoIn Section  show that this aspect is currently limited by the need for anticipating both the function class and the necessary intervention value of the environment we want to apply the learned policy on.

\subsection{POMDP Formulation of Causal Discovery Through Interventions}\label{sec:formulation}

To conveniently model the causal discovery process where there is partial observability, we will use a POMDP. However, since such a formalization is not entirely obvious, it constitutes our first contribution, which we will present in this section. 

\paragraph{State Space: } Our environment is determined by SCMs.\footnote{Our approach can be applied to any data-generating process that allows for sampling from and intervening on its variables.} Therefore, each state will correspond to an SCM. 
%Since we will assume no exogenous variables throughout the paper,
We shall describe each SCM as the set of functions that determine the endogenous variables. Therefore, having $n$ endogenous variables, a state is a set $s=\{f_0, ..., f_{n-1}\}$ of functions that define the current SCM. Furthermore, each state contains the ground truth observational graph $G^*_s$ induced by the observational SCM $M_{do(\emptyset)}$.%variables $\{X_0, ... X_{n-1}\}$.

\paragraph{Action Space: } We model our action space as a multi-discrete space $A=~[n +1]~\times~[2n(n-1)+1]$ where the notation $[n] = \{1,2,\ldots,n\}$ with $n$ being the number of nodes in the graph. The first dimension of the action space represents the endogenous variables of the current SCM and determines the \emph{intervention targets}. For each variable $X_i \in \mathcal{X}$, there is an action $do(X_i = c)$, where $c$ is a predefined constant. In addition, the agent can do \emph{nothing} and just collect observational data. In total, this dimension of the action space has $n+1$ elements. The second dimension represents the \emph{structural actions}. Each action in this space indexes the removal and addition of edges on the currently estimated graph. Additionally, the agent can perform a \emph{void} structural update %can be skipped 
which leads to a total size of $2n(n-1)+1$ actions in this dimension since we disallow reflective edges. The action space scales quadratically with the number of nodes. To address this problem, we mask the possible actions at every step such that the agent cannot add an already existing edge or remove a non-existing edge. This effectively halves the size of this action space dimension.

\paragraph{Transition Dynamics: } Each episode starts in the observational SCM $M_{do(\emptyset)}$ where no intervention is performed. An intervention $do(X_i)$ on this SCM deterministically leads to a new SCM $M_{do(X_i)}$ where $f_i$ is replaced by some constant $c$. This effectively replaces $f_i$ in the state. With an intervention $do(X_j)$, we transition from $M_{do(X_i)}$ to $M_{do(X_j)}$, that is, $T(M_{do(X_i)}, do(X_j), M_{do(X_j)}) = 1$ or, equivalently, $T(M_{do(X_i)}, do(X_j)) = M_{do(X_j)}$. A minimal example of the transition dynamics of our approach can be seen in Figure \ref{fig:overview}.

\paragraph{Observations: } At each step $t$, the agent collects the value of the endogenous variables $\{x_0, \ldots, x_{n-1}\}$ from the joint distribution $P_{M_{do(X)}}(\mathcal{X})$ induced by the current SCM $M_{do(X)}$. Therefore, the observation is $o_t \sim P_{M_{do(X)}}(\mathcal{X})$. 

\paragraph{State Representation:} The use of a single observation $o_t$ is not sufficient to determine the best action in POMDPs. Thus, the agent has to build its own state representation $h_t$ using the history of observations and actions \cite{Sutton2018ReinforcementIntroduction}. We denote the history of observations and actions by $h_t=[x_0, a_0, ..., x_t, a_t]$.% to build its \textit{internal} state representation $z_t$.
%\footnote{Note that the currently estimated graph is fully determined by the history of actions.}

% We are in POMDP so the agent has to build its own state representation based on history information. The history, input of value function and policy, is composed of the sequence of observations from the environments, of actions, and causal graphs estimates.

\paragraph{Reward: } The structural Hamming distance (SHD) measures the distance between two DAGs by counting the number of different edges. Since our goal is to minimize the distance between the generated and the ground truth observational graphs at every step, we consider the SHD as a natural candidate for our reward function. For a ground truth observational graph $G^*_s$, a graph estimate $\hat{G}_t$, and a graph estimate $\hat{G}_{t'}$ in the consecutive step $t'$, we define the potential-based reward \cite{Jenner2022CalculusGradient} as $r(s, a) = SHD(G^*_s, \hat{G}_t) - SHD(G^*_s, \hat{G}_{t'})$. 

To simplify, we rewrite our reward function in the following way: Let $E(a)$ be the directed edge that is manipulated in action $a$. Then, when adding an edge $E(a)$, our reward becomes:
\begin{equation}\label{eq:reward}
    r(s, a) = \begin{cases}
    1 & \text{if } E(a) \in G_s^*\\
    -1 & \text{if } E(a) \not\in G_s^*\\
    %0 & \text{if } E(a) = \emptyset
    \end{cases}
\end{equation}
When removing an edge $E(a)$ our reward becomes $-r(s,a)$. In all other cases ($E(a)=\emptyset$) the reward is $0$. 

This formulation has the computational advantage that only $E(a)$ must be compared to the edges of $G_s^*$ instead of comparing the entire graph $\hat{G}_{s'}$. Furthermore, it makes the reward denser and depends only on $s$ and $a$ instead of relying on the entire history of structural actions that make up the current graph estimate. In Appendix \ref{apx:reward} we demonstrate that this formulation is equivalent to $SHD(G^*_s, \hat{G}_s) - SHD(G^*_s, \hat{G}_{s'})$.

\subsection{Data-Generation}\label{sec:data-generation}
We train our CORE agents using a training set of DAGs. In addition, we have an evaluation set of DAGs that the agent has not seen during training. To ensure that the evaluation set does not include any graphs from the training set, we first create a set of unique DAGs, shuffle it to ensure equal sparsity throughout the list, and then divide it into training and evaluation sets. As per common assumption in machine learning, we assume that having more graphs in the training set will help us to generalize better to the evaluation set.

Since the space of DAGs grows superexponentially in the number of its nodes, it quickly becomes infeasible to generate all possible graph structures with $n$ nodes. For this reason, we generate all possible graphs only for graphs with 3 nodes (for a total of 25 graphs) and graphs with 4 nodes (for a total of 543 graphs). For graphs with more than 4 nodes, we generate subsets of the possible graphs. Similarly to many works in the literature, each graph is generated as an Erd\"{o}s-R\'{e}nyi \cite{Erdos1959OnGraphs} graph with an edge probability of 0.2. We diversify the training data by generating SCMs based on these graphs by sampling a function $f_i(Pa_G(X_i))$ from a class of possible functions for every node $X_i$ in the graph $G$ at the beginning of each episode.

\begin{table*}[h!]
\begin{tabular}{lccccc}
\hline
       & 3 variables     & 4 variables     & 5 variables      & 8 variables     & 10 variables   \\ \hline
random & $3.29 \pm 0.97$ & $5.92 \pm 1.41$ & $8.33 \pm 1.22$  &$11.08 \pm 2.80$ &$14.85 \pm 4.01$\\
empty  & $1.80 \pm 0.90$   & $3.80 \pm 1.10$   & $6.20 \pm 0.42$  &$5.10 \pm 1.60$    &$7.0 \pm 2.50$   \\
MCD    & $1.85 \pm 1.10$ & $4.97 \pm 1.61$ &$6.18 \pm 0.44$  &           -     &         -      \\
CORE   & $0.50 \pm 0.50$ & $0.54 \pm 0.65$ &  $1.26 \pm 1.06$  &$2.04 \pm 1.64$& $5.16 \pm 2.69$\\ \hline
\end{tabular}
\caption{Average SHDs on the test set of SCMs with unseen causal structures.}
\label{tab:generalization_results}
\end{table*}

\subsection{Learning Approach}
The CORE agent is based on the DQN algorithm \cite{Mnih2015Human-levelLearning}, but it maintains two multilayer perceptron Q networks simultaneously. One network, $Q_{st}(h, a_{st}; \Theta_{st})$, estimates the Q-values specific to the structural updates, the other $Q_{in}(h, a_{in}; \Theta_{in})$ maintains the values for the interventions. Each of the two networks comes with its own target network, and the loss is computed separately. Note that Q values are determined on the basis of the history of observations, as is often done when applying DQN to POMDP problems \cite{Mnih2015Human-levelLearning}. The networks are identical except for the output layers due to different dimensionalities of the action space. The overall output of the Q-function is the concatenation of the individual Q-values, and our greedy policy picks $a_{in}$ and $a_{st}$ in such a way that the corresponding Q-values are maximized. %$Q(h, a; \Theta_{in}, \Theta_{st}) = (Q_{in}(h, a_{in}; \Theta_{in}), Q_{st}(h, a_{st}; \Theta_{st}))$ with $a=(a_{in}, a_{st})$. 

At the procedural level, our algorithm generates a new SCM based on a random sample of the graph data at the beginning of each episode. Furthermore, we start every episode with an empty graph estimate. The inference time for a given SCM is fully determined by the fixed number of steps per episode. %All structure updates and interventions for a given SCM have to be performed within an episode of fixed length. 
This puts a hard upper bound on the number of samples and makes inference highly efficient. Figure \ref{fig:overview} gives an overview of our agent. 

\section{Generalization to Unseen Structures}\label{sec:generalization}
In this section, we empirically validate whether our learned policy constitutes a good causal discovery algorithm. To this end, we train our model on a training set of SCMs with known causal structures and evaluate it on SCMs with causal structures that were not seen during training. 

\subsection{Training Data}\label{sec:generalization_data}
For this experiment, we generate graphs with 3, 4, 5, 8, and 10 variables as described in Section \ref{sec:data-generation}. We split the generated graphs into training and test sets as follows: We first generate the graphs (25, 543, 16000, 91000, 101000), and then split the final list into train and test sets with splits 19/6, 401/142, 15000/1000, 90000/1000, 100000/1000 for 3, 4, 5, 8 and 10 variables, respectively. We limit the number of test graphs to 1000 since the evaluation would otherwise slow down the training prohibitively.

At the beginning of each episode, a graph is sampled from the data set. To generate the SCMs in accordance with Section \ref{sec:data-generation}, we define a class of linear additive functions. For each node $X_i$ in the graph $G$, we sample $\mid Pa_G(X_i) \mid$ weights from $Uniform(0.5, 2.0)$. The generated function for this node is then $X_i \leftarrow \Sigma_{X_j\in Pa(X_i)} w_j \cdot x_j$ for the current values $x_j$ of the parents of $X_i$. If $X_i$ is a root node, we assign a default value of 0. We use an intervention value of 20 to provide a strong signal about the causal structure w.r.t. to the true causal effect sizes. This value can be considered as a hyperparameter and we discuss its impact further in Section \ref{sec:real_world}.

\subsection{Experimental Setup}
We evaluate the generalization capability of CORE w.r.t. the available baselines. While AVICI \cite{Lorch2022AmortizedLearning} learns a graph generator that estimates the causal structure of an offline dataset, CORE operates in a few-shot online data sample regime. Therefore, a meaningful comparison with AVICI is out of reach. Consequently,  MCD \cite{Sauter2023ADiscovery} is, to the best of our knowledge, the only SOTA method that learns a CD algorithm that actively intervenes. Furthermore, we compare with the \emph{random} baseline that generates random DAGs, and the \emph{empty} baseline which represents the empty graph. 

We train both MCD and our approach for the same amount of steps and align all relevant hyperparameters including the neural network sizes. MCD runs into difficulties when scaling up to graphs with more than 4 nodes. Due to such computational infeasibility, we cannot run experiments for MCD on SCMs with 8 or 10 variables.  A detailed description of the hyperparameters and the architectures used can be found in Appendix \ref{apx:hyperparameters}. We set a maximum compute budget via a timeout of 25 training hours. The precise hardware configuration can be found in Appendix \ref{apx:hardware_requirements}. We paid special attention to setting comparable episode lengths for both approaches. For the sake of fair comparison, we set the episode length of our approach to half the episode length used in MCD,  since our approach can perform interventions and structure updates synchronously, while MCD can only perform them sequentially. 

\subsection{Results}
\begin{figure*}[h!]
    \centering
    \includegraphics[width=0.99\textwidth]{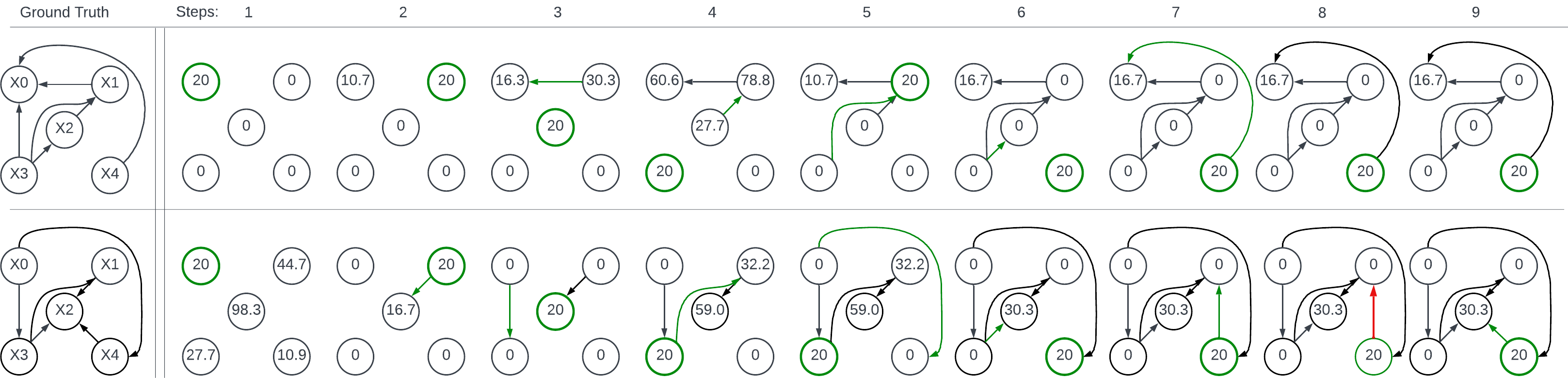}
    \caption{Two examples of how the learned CORE policy estimates the causal structure of two unseen SCMs described in Equations \eqref{eq:scm1} and \eqref{eq:scm2}. Green elements indicate intervention (do (c = 20)) and structural update (adding an edge) in the current step, respectively. The red arrow indicates the deletion of an edge.}
    \label{fig:example}
\end{figure*}

The results in Table \ref{tab:generalization_results} are from applying the learned models to three SCMs randomly generated for each graph in the held-out test set. The model that achieved the best performance during training was chosen for each evaluation.

We can see the favorable performance of CORE compared to MCD and the trivial baselines. For all sizes of graphs tested, we observe that our approach estimates graphs that are closer to the ground-truth structures than the other approaches in less than 34 milliseconds per graph. We generate an average of 0.5, 0.5, 1.3, 2.0, and 5.2 wrong edges in graphs with 3, 4, 5, 8, and 10 nodes, respectively. Even in a set of graphs with 10 variables, our approach adds approximately 2 correct edges out of 90 potential edges while only observing 15 data points. %\footnote{The number of samples is determined by the episode length described in Appendix~\ref{apx:hyperparameters}.} 
For smaller graphs, the ratio of correctly identified edges is even higher.

%We attribute the improvement over the MCD to a variety of aspects. Firstly, modeling the structural actions and the intervention actions with separate networks makes it easier to learn the corresponding Q-functions. We argue that this is because it allows the two networks to model distinct representations that might need to be different for structural and intervention actions.
%Secondly, representing the reward densely instead of a summary at the end of each episode often improves performance. 

We attribute the improvement over MCD to a variety of aspects. First, addressing the structural actions and the intervention actions with separate networks makes learning the corresponding Q-functions more efficient. This is because two separate networks have a greater degree of freedom to represent structural and intervention actions that are inherently different. Second, representing the reward densely instead of a summary at the end of each episode often improves performance \cite{Ng1999PolicyShaping}. 
Third, instead of learning to integrate observations from the environment via a long short-term memory (LSTM) \cite{Hochreiter1997LongMemory}, we directly input the history of samples into our policy.
And, lastly, by not generating graphs at runtime, but rather having a pre-defined training set, we avoid a significant computational overhead.% This aspect also allows to model more sophisticated action spaces, such as latent action spaces REFS, in extensions of this work. 

Furthermore, we observe the rather unfavorable performance of MCD compared to the baselines. We partly attribute this to a lack of extensive hyperparameter tuning, since this would likely have been needed to achieve the results in \cite{Sauter2023ADiscovery}. For the 4 and 5 variable cases, MCD reached the timeout of 25 hours.

Given these results, we conclude that CORE is capable of successfully learning a CD algorithm that can be applied to previously unseen causal structures. Even for these cases, our approach estimates the ground truth graph accurately without having to retrain on the new structure. Furthermore, we show that with CORE's novelties, we are able to scale towards graph sizes of more relevance for real-world applications, while simultaneously increasing training efficiency.

\subsection{Examples} \label{sec:scm_example}
We present qualitative results on how our learned policy performs on the following two randomly selected example SCMs with unseen causal structures:

\begin{equation}\label{eq:scm1}
    M_{do(\emptyset)}^0 = 
    \begin{cases}
        X_0 \leftarrow 0.54 \cdot X_1 + 0.91\cdot X_3 + 0.83 \cdot X_4,\\
        X_1 \leftarrow 1.52\cdot X_2 + 1.84\cdot X_3,\\
        X_2\leftarrow 1.38 \cdot X_3,\\
        X_3\leftarrow 0, \quad X_4 \leftarrow 0\\
    \end{cases}
\end{equation}
\begin{equation}\label{eq:scm2}
M_{do(\emptyset)}^1 =
     \begin{cases}
        X_0 \leftarrow 0,\\
        X_1 \leftarrow 1.61\cdot X_3,\\
        X_2\leftarrow 0.83\cdot X_1 + 1.60\cdot X_3 + 1.5\cdot X_4,\\
        X_3\leftarrow1.39\cdot X_0,\\
        X_4 \leftarrow 0.54 \cdot X_0,\\
    \end{cases}
\end{equation}

In Figure \ref{fig:example}, we can see that in both cases the agent identifies the underlying causal structure almost correctly, with the exception of the missing edge $X_3 \rightarrow X_0$ in the first instance. In the second instance, it even recognizes an error and corrects it in Step 7. 

\section{On the Importance of Jointly Learning an Intervention Policy}
As described in Section \ref{sec:setup}, our method is designed to jointly learn a causal graph generator and an intervention policy. In this section, we show that learning an intervention policy, aimed at performing the interventions that are most informative for CD, helps in learning a CD policy.

It is worth noting that our agent does not receive any specific reward that represents the quality of the intervention performed in the environment. Instead, the reward function depends only on the structural update of the currently estimated causal structure (see Equation \eqref{eq:reward}). Since, for the full identification of the causal structure, interventions are generally needed \cite{Bareinboim2022OnInference}, our agent has to learn to perform interventions to update the estimate of the causal structure. Therefore, our agent receives good rewards only if it performs interventions that are relevant to discover the current causal structure.

Although it is clear that interventions, in general, are helpful for CD, we argue that learning an intervention policy by measuring the usefulness for structure identification helps the overall learning process. Especially when the budget for performing interventions is very restrictive, as is the case in many real-world applications,  it is crucial to perform the interventions that are most informative about the underlying causal structure \cite{Shanmugam2015LearningInterventions, Tigas2022InterventionsScale}. Which interventions are the most informative ones depends on the causal structure that is currently being discovered. This further motivates learning the intervention policy jointly with the structure generation policy. In this section, we empirically show that there is in fact a benefit in learning an intervention policy jointly with the CD policy. 

\subsection{Experimental Setup}
To show the hypothesized performance gain, we compare the performance of the agent that learns the intervention policy and the structure generation policy jointly, to an agent that learns only the structure generation policy and randomly picks an intervention target at each step. We train the two agents 3 times each on environments with 4 endogenous variables. The graphs, function classes, and hyperparameters remain the same as described in Section \ref{sec:generalization_data}.

\subsection{Results}
\begin{figure}
    \centering
    \includegraphics[width=0.4\textwidth]{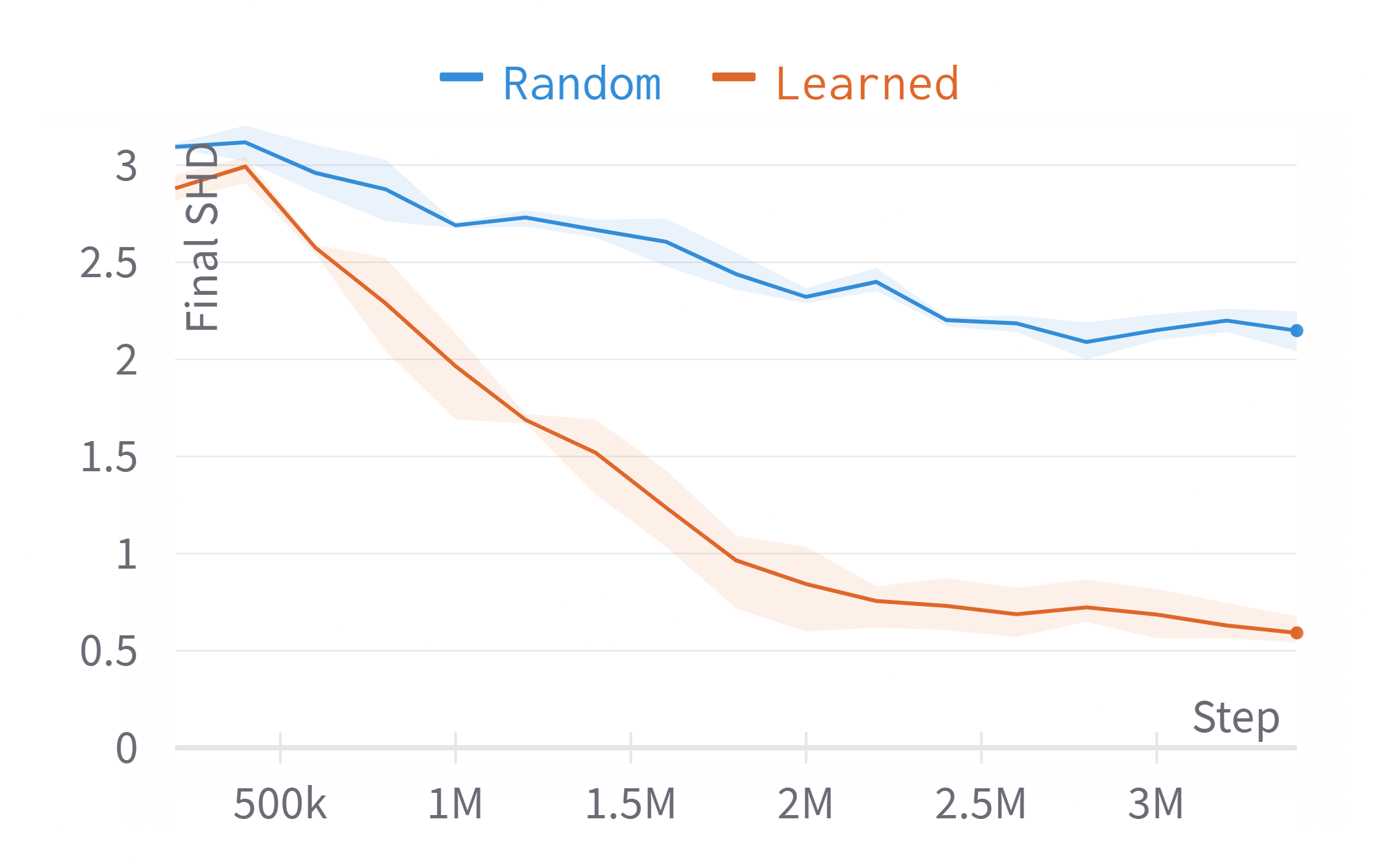}
    \caption{Plot of the average SHD on the test set (lower is better). We present the means over three training runs of CORE with random interventions (blue) and when jointly learning an intervention policy (red) over graphs with 4 variables.}
    \label{fig:rand_vs_learned}
\end{figure}

Figure \ref{fig:rand_vs_learned} shows the aggregated results for the two types of agents.
We can see that learning the intervention policy jointly with the CD policy results in a significantly better estimation of the ground truth causal structure. Additionally, it decreases the number of learning steps needed to reach a certain level of performance. Lastly, Figure~\ref{fig:rand_vs_learned} suggests that even with random interventions, our approach performs reasonably well (average SHD of $\sim2.3$). This indicates the robustness of the CORE agent to less informative interventions. 

Overall, we observe that the ability to learn the intervention policy is an integral part of learning a CD policy and that rewarding interventions leading to better structure estimates is sensible.

\section{Applicability to the Real-World}\label{sec:real_world}
\begin{table*}[h]
\begin{tabular}{l|ccc|ccc|ccc}
\hline
      & \multicolumn{3}{c|}{3 Variables}         & \multicolumn{3}{c|}{4 Variables}         & \multicolumn{3}{c}{5 Variables}          \\
      & lin & lin + noise  & interaction         & lin & lin + noise  & interaction   & lin & lin + noise  & interaction          \\ \hline
empty & \multicolumn{3}{c|}{$1.8\pm0.9$}         & \multicolumn{3}{c|}{$3.8\pm1.1$}         & \multicolumn{3}{c}{$6.2 \pm 0.4$}        \\ \hline
lin 5    & $0.2\pm0.4$  & $1.2\pm0.4$ & $2.2\pm0.7$ & $0.6\pm0.7$  & $0.8\pm0.8$ & $3.4\pm1.9$ & $1.4\pm1.1$  & $1.8\pm1.2$ & $6.1\pm2.4$ \\
lin 20 & $0.5 \pm 0.5$ & $0.8\pm0.4$ & $2.3\pm0.5$ & $0.5\pm0.7$ & $0.6\pm0.6$ & $3.8\pm1.8$ & $1.3\pm1.1$ & $1.2\pm1.0$ & $6.5\pm2.2$ \\
lin+noise 5   & $0.7\pm0.7$  & $0.6\pm0.7$ & $1.3\pm0.7$ & $0.6\pm0.7$  & $0.8\pm0.8$ & $3.6\pm2.0$ & $2.0\pm1.3$  & $2.3\pm1.3$ & $5.9\pm2.1$ \\
lin+noise 20  & $0.2\pm0.4$  & $0.8\pm0.4$ & $2.0\pm0.8$ & $0.6\pm0.6$  & $0.7\pm0.7$ & $4.3\pm1.7$ & $1.2\pm1.0$  & $1.2\pm1.0$ & $6.8\pm2.2$ \\
interaction 5  & $0.8\pm0.7$  & $1.3\pm0.7$ & $1.0\pm0.5$ & $4.1\pm1.3$  & $4.1\pm1.3$ & $4.0\pm1.3$ & $7.9\pm1.7$  & $7.8\pm1.7$ & $7.9\pm1.7$ \\ \hline
\end{tabular}
\caption{We show the performance of trained CORE policies on unseen graphs with various function classes for their corresponding SCMs. Each row describes which function class the model was trained on and what intervention value it uses. Each column describes the function class it was tested on. Empty describes the baseline that generates the empty graph.}
\label{tab:limitations_results}
\end{table*}
%\section{Learning Causal Discovery Algorithms for Real-World}
Throughout this work, we acknowledge the capability of learning CD algorithms that can be applied to environments with previously unseen causal structures with up to 10 variables. This constitutes a substantial improvement over the SOTA when it comes to the application of learned CD algorithms in the real world, as many problems can be modeled with ~10 variables \cite{Zwitter1988BreastCancer, Sachs2005CausalData}. Therefore CORE, %as the first learned CD algorithm (i.e., general policy that approximates an algorithm), 
reaches graph sizes that are relevant in CD. However, we acknowledge that many applications with up to 5000 variables \cite{Maathuis2010PredictingData} are currently out of reach.  In this section, we shed light on some of the limitations that this approach currently has with regard to applying it in a real-world scenario. 

%Here, we specifically investigate some of these limitations that are due to two design aspects that are interconnected in more detail.

Here, we specifically investigate two design aspects that limit real-world applicability, and they are interconnected.
First, during training, the functions of the SCMs are sampled from a specific function class, and for some function classes (e.g., non-linear functions), discovering the true causal structure can be harder than for others \cite{Glymour19Review}. Consequently, as we will show in this section, learning a CD algorithm for these classes of functions is more difficult.

Second, our approach is tailored to generate graph estimates for the function class on which it was trained. This means that when used for different causal structures, the same function class is expected during inference. Consequently, this can lead to a decrease in performance if the function class is altered.  We expect an exception for this for function classes that are either very similar to the training functions or that subsume them. Therefore, when CORE is trained with the intention of being used in a real-world setting, the real-world function class has to be anticipated during training.

\subsection{Transferability across Noise and Non-Linearity}
Motivated by these aspects, we show the difficulty in training CD policies on some function classes and test CORE on function classes that it was not trained on.

\subsubsection{Experimental Setup}
For the data-generating processes in this section, we test how noise and non-linearities influence the performance and transferability of CORE. Therefore, we use three function classes that define each function class $f_i(Pa(X_i), U_i)$ in an SCM as follows:
\begin{itemize}
    \item \emph{linear: }  $f_i = \Sigma_{X_j\in Pa(X_i)} w_j \cdot x_j$ (same as Section \ref{sec:generalization_data})
    \item \emph{linear + noise: } $f_i = \Sigma_{X_j\in Pa(X_i)}  w_j \cdot x_j + u_i$, where $u_i \sim \mathcal{N}(0,0.5)$
    \item \emph{interaction: } $f_i = \Sigma_{X_j\in Pa(X_i)} w_j \cdot x_j + x_k \cdot x_l$, where $X_j, X_k \in Pa(X_i)$ 
\end{itemize}
where the lowercase $x_i$ represents the current value of the variable $X_i$. Whenever a node is a root node, we set a default value of 0.

We train two CORE models for various graph sizes for each of the two linear functions, one with an intervention value of 5 and the other with an intervention value of 20. This setup gives us further insight into how the signal-to-noise ratio affects our performance. For the interaction function, we train one model with an intervention value of 5. We then tested all the trained models in all three function classes.

\subsubsection{Results}
In Table \ref{tab:limitations_results}, we show how models that were trained with one function class perform when applied to various function classes with previously unseen causal structures. 

% doesn't work well when applied to different function class
Our first observation is that, as hypothesized, the application of our learned CD algorithm on previously unseen function classes is problematic if the testing function class is very different from the training function class. While applying the linear models to their noisy/non-noisy counterparts still leads to good estimates, applying them to the interaction data mostly fails. 
%This strengthens our argument that the function class chosen for training must be informative about the function class that is encountered during testing.
This supports our claim that the function class chosen for training must be informative about the function class encountered during testing.

% trained on more complex function is hard, but will work on easier function
Looking at the model that was trained on interaction data, we observe two interesting aspects. First, for graphs with 4 and 5 variables, CORE fails to learn a CD policy that generalizes to unseen graphs. We believe that, given the right hyperparameters and a sufficient training budget, we can solve the task for larger graphs, based on the results obtained from smaller graphs. However, these results demonstrate that the performance of a learned CD algorithm depends on the complexity of the SCM function that generates the data. Second, we observe that, for the interaction case, the learned policy can be successfully applied to the linear function. We argue that this is because the interaction data encompasses the linear data as well. This suggests that if the function classes used during training are broad enough, the CD algorithm that is learned will be relatively more applicable to real-world scenarios.

% c=20 vs c=5
When comparing linear models with a higher intervention value with those with a lower intervention value, we observe that they tend to perform better on function classes that they successfully learned. We attribute this to a higher signal-to-noise ratio w.r.t. the data-generating process.

% Conclusion
Overall, we can say that learning CD algorithms is limited by the function class that is observed during training. This currently obstructs their application to real-world scenarios, but finding more general functions on which CORE can be trained is a promising research direction.

\subsection{Further Limitations}
Apart from aspects related to the function classes on which CORE is trained, we point out the additional limitations of learning a model that is applicable to the real world.

One of them is the assumption of being able to intervene on any variable with any target value. In real-world scenarios, it might be that some variables cannot be manipulated (imagine changing the outside temperature) or a good target value is unknown during training. Furthermore, CORE-like algorithms might suffer from the presence of unobserved confounders. For both the unknown target value and unobserved confounders, there is hope that augmenting the learning procedure in the future will overcome these limitations.

\section{Summary and Conclusion}

In this paper, we introduce CORE, a deep RL-based approach to tackle the task of causal discovery. CORE learns a policy to sequentially perform informative interventions and generate candidate causal graphs from scratch. Moreover, the learned policy generalizes to previously unseen graphs of up to 10 variables in size. CORE outperforms the current SOTA baseline (i.e., MCD \cite{Sauter2023ADiscovery}) both in the number of variables it can deal with and in the accuracy of the estimated structure. Furthermore, it demonstrates that by learning to perform the most informative interventions, highly sample-efficient CD algorithms can be learned ($\sim15$ data samples for 10 variables).  

Such improvement can be attributed to several key design features. One such feature is the imposed additional structure in the policy that separates the networks for interventions and structural updates. However, such separation is not completely isolated. As shown in our ablation study, CORE learns to perform relevant interventions outperforming random interventions. Learning which interventions are relevant is guided solely through a dense reward that assesses the accuracy of the generated graph. %Performing relevant interventions  is guided through another key feature  that is the representation of the the reward at each step instead of a summary at the end of each episode. 

Moreover, we outlined the real-world applicability of our approach in terms of the number of variables and generalizability across more complex function classes. For the former, while the number of variables that CORE can deal with matches some domains, for some other domains, usual practice is still out of reach.  For the latter, it turns out that CORE delivers good estimates when it comes to linear functions, training on noisy functions, and testing on non-noisy counterparts, and vice versa. However, generalizing to more complex classes such as non-linear functions necessitates improvements.  Furthermore, we empirically confirmed that training on more complex function classes and testing on simpler classes yields promising estimates. Such an observation can serve as a key idea along the road of applying these methods to real-world problems.  

Future work includes investigating the limitations of our approach in the presence of confounders, soft interventions, and determining the right target value for interventions. In particular, it would be interesting to understand how robust our approach is when it comes to different topologies regarding such non-intervenable variables and unobservable confounders. Another avenue that we plan to address is an extensive study of the transferability of our approach across different function classes, such that learned CD algorithms that autonomously plan interventions can be applied to real-world data. 

% \bibliographystyle{unsrtnat}
% \bibliography{references,referencesandreas}

%%%%%%%%%%%%%%%%%%%%%%%%%%%%%%%%%%%%%%%%%%%%%%%%%%%%%%%%%%%%%%%%%%%%%%%%

%%% The acknowledgments section is defined using the "acks" environment
%%% (rather than an unnumbered section). The use of this environment 
%%% ensures the proper identification of the section in the article 
%%% metadata as well as the consistent spelling of the heading.

\begin{acks}
We thank Frank van Harmelen for his valuable input throughout this project and the anonymous reviews for helping to improve the final version of this work.
This research was partially funded by the Hybrid Intelligence Center, a 10-year programme funded by the Dutch Ministry of Education, Culture and Science through the Netherlands Organisation for Scientific Research, \url{https://hybrid-intelligence-centre.nl}, grant number 024.004.022
\end{acks}

%%%%%%%%%%%%%%%%%%%%%%%%%%%%%%%%%%%%%%%%%%%%%%%%%%%%%%%%%%%%%%%%%%%%%%%%

%%% The next two lines define, first, the bibliography style to be 
%%% applied, and, second, the bibliography file to be used.

\bibliographystyle{ACM-Reference-Format} 
%\bibliography{sample}
\bibliography{main}

%%%%%%%%%%%%%%%%%%%%%%%%%%%%%%%%%%%%%%%%%%%%%%%%%%%%%%%%%%%%%%%%%%%%%%%%
\appendix
\mbox{~}
\clearpage

\newpage

\section{Equivalence of our Reward and the Difference in SHDs}\label{apx:reward}

Let $G^*_s$ be the observational ground-truth graph of the current SCM $M^i_{do(\emptyset)}$. Furthermore, let $\hat{G}_{s}$ be the estimated graph at state $s$, $\hat{G}_{s'}$ the estimated graph in the consecutive state and $E(a)$ the edge that is manipulated by action $a$. We show that 

\begin{equation}\label{eq:shd_add}
SHD(G^*_s, \hat{G}_{s})-SHD(G^*_s, \hat{G}_{s'}) = \begin{cases}
    1 & if E(a) \in G^*_s\\
    -1 & if E(a) \not\in G^*_s\\
    0 & if E(a) = \emptyset
    \end{cases}
\end{equation}
\\if $a$ adds an edge and 

\begin{equation}\label{eq:shd_minus}
SHD(G^*_s, \hat{G}_{s})-SHD(G^*_s, \hat{G}_{s'}) = \begin{cases}
    -1 & if E(a) \in G^*_s\\
    1 & if E(a) \not\in G^*_s\\
    0 & if E(a) = \emptyset
    \end{cases}
\end{equation}
\\if $a$ deletes an edge. In the following, we derive this equality. For a simplified notation, we write $E_G$ for the set of edges in $G$. We start by decomposing the difference in its set-theoretic components.
    \begin{align*}
        SHD(G^*_s, \hat{G}_{s})-SHD(G^*_s, \hat{G}_{s'}) &= \mid E_{G^*_s}  \setminus  E_{\hat{G}_{s}} \mid + \mid E_{\hat{G}_{s}} \setminus  E_{G^*_s} \mid \\&- \mid E_{G^*_s}  \setminus  E_{\hat{G}_{s'}} \mid - \mid E_{\hat{G}_{s'}} \setminus  E_{G^*_s} \mid
    \end{align*}

We now distinguish the two cases of adding and deleting, and edge. In these cases, the terms become:

\paragraph{Case 1; an edge is \emph{added} to $\hat{G}_s$: } Then 
\begin{equation}\label{eq:shd_minus_first}
    \mid E_{G^*_s}  \setminus  E_{\hat{G}_{s}} \mid -\mid E_{G^*_s}  \setminus  E_{\hat{G}_{s'}} \mid = \begin{cases}
    1 & if e \in E_{G^*_s}\\
    0 & if e \not \in E_{G^*_s}
\end{cases}
\end{equation}
and 
\begin{equation}\label{eq:shd_minus_second}
    \mid E_{\hat{G}_{s}} \setminus  E_{G^*_s} \mid - \mid E_{\hat{G}_{s'}} \setminus  E_{G^*_s} \mid = \begin{cases}
    0 & if e \in E_{G^*_s}\\
    -1 & if e \not \in E_{G^*_s}
\end{cases}
\end{equation}

Finally, considering the case where the edge was not manipulated ($E(a)=\emptyset$), we add Equations \eqref{eq:shd_minus_first} and \eqref{eq:shd_minus_second} and arrive at Equation \eqref{eq:shd_add}. 

\paragraph{Case 2; an edge is \emph{removed} from $\hat{G}_s$: } In this case, the results in Equation \eqref{eq:shd_minus_first} and \eqref{eq:shd_minus_second} are multiplied by $-1$. After again adding the inverted parts, we arrive at Equation \eqref{eq:shd_minus}.

\section{Hyperparameters}\label{apx:hyperparameters}
\begin{table*}[t!]
\resizebox{\textwidth}{!}{
\begin{tabular}{llcccccc}
                              &      & shared layers & shared layers size & policy layers & policy layers size & episode length & total training steps \\ \hline
3 variables  & MCD  &     1 LSTM      &        64            &       2        &        128            &       10         &     2000000                 \\ 
                              & CORE &         0      &           -         &         3      &           128         &           5     &      2000000                \\ \hline
4 variables  & MCD  &        1 LSTM  &          64          &       2        &         128         &        16        &     3500000                 \\
                              & CORE &         0      &           -         &         3      &           128         &         8       &      3500000               \\ \hline
5 variables  & MCD  &       1 LSTM   &           128         &         2      &           256         &        20        &      4500000                 \\
                              & CORE &         0      &             -       &       3       &            256        &        10        &      4500000                 \\ \hline
8 variables  & MCD  &        -       &         -           &         -      &     -               &        -        &          -            \\
                              & CORE &           0    &            -        &      2         &         1024           &        12        &        45000000              \\ \hline
10 variables & MCD  &      -         &           -         &        -       &     -               &        -       &             -         \\
                              & CORE &        0       &           -         &     3          &        1024            &         15       & 90000000    \\  \hline               
\end{tabular}
}
\caption{The hyperparameters that we used to obtain our results. These parameters are the same for all experiments we performed in this paper.}
\label{tab:hyperparameters}
\end{table*}
Table \ref{tab:hyperparameters} describes the main hyperparameters that we used throughout this paper. For MCD \cite{Sauter2023ADiscovery}, we additionally used the default hyperparameters that were used in the original paper.

\section{Hardware Requirements}\label{apx:hardware_requirements}
We ran training and evaluation on the DAS-6 compute cluster \citep{Bal2016ATerm}. At inference time, CORE generates a graph estimate in about 18-34 milliseconds, depending on the size of the policy network and the episode length. 

For training, CORE took approximately 51min, 1h40min, 2h5min, 23h, 30h to train for 3, 4, 5, 8, 10 variables, respectively. For the 3, 4, and 5 variable case these are the results on a 24 core machine with an RTX4000 GPU, for 8 and 10 variables the results are on a 48 core A100 machine.

\end{document}

%%%%%%%%%%%%%%%%%%%%%%%%%%%%%%%%%%%%%%%%%%%%%%%%%%%%%%%%%%%%%%%%%%%%%%%%